\documentclass{article}

\usepackage{microtype}
\usepackage{graphicx}
\usepackage{subcaption}
\usepackage{booktabs} 

\usepackage{tikz}
\usepackage{pgfplots}

\usepackage{hyperref}
\usepackage{enumitem}

\usepackage[accepted]{mlsys2020}

\usepackage[algo2e,ruled,vlined]{algorithm2e}
\mlsystitlerunning{On-device Federated Learning with Flower}

\begin{document}

\twocolumn[
\mlsystitle{On-device federated learning with Flower}


\mlsyssetsymbol{equal}{*}

\begin{mlsysauthorlist}
\mlsysauthor{Akhil Mathur}{cam,nokia}
\mlsysauthor{Daniel J. Beutel}{cam,adap}
\mlsysauthor{Pedro Porto Buarque de Gusm\~{a}o}{cam}
\mlsysauthor{Javier Fernandez-Marques}{ox}
\mlsysauthor{Taner Topal}{cam,adap}
\mlsysauthor{Xinchi Qiu}{cam}
\mlsysauthor{Titouan Parcollet}{au}
\mlsysauthor{Yan Gao}{cam}
\mlsysauthor{Nicholas D. Lane}{cam}
\end{mlsysauthorlist}

\mlsysaffiliation{cam}{University of Cambridge, UK}
\mlsysaffiliation{adap}{Adap, Germany}
\mlsysaffiliation{nokia}{Nokia Bell Labs, UK}
\mlsysaffiliation{au}{Avignon Université, France}
\mlsysaffiliation{ox}{University of Oxford}

\mlsyscorrespondingauthor{Akhil Mathur}{akhilmathurs@gmail.com}


\vskip 0.2in


\begin{abstract}
Federated Learning (FL) allows edge devices to collaboratively learn a shared prediction model while keeping their training data on the device, thereby decoupling the ability to do machine learning from the need to store data in the cloud. Despite the algorithmic advancements in FL, the support for on-device training of FL algorithms on edge devices remains poor. In this paper, we present an exploration of on-device FL on various smartphones and embedded devices using the Flower framework. We also evaluate the system costs of on-device FL and discuss how this quantification could be used to design more efficient FL algorithms.

\end{abstract}
]

\printAffiliationsAndNotice{}  

\section{Introduction}

We have seen remarkable progress in enabling the execution of deep learning models on mobile and embedded devices to infer user contexts and behaviors~\cite{warden2019tinyml, 10.5555/3327144.3327315,DBLP:journals/corr/abs-1906-05721,malekzadeh2019privacypreserving,10.1145/3300061.3345447}.  This has been powered by the increasing computational abilities of edge devices as well as novel software optimizations to enable cloud-scale models to run on resource-constrained devices. However, when it comes to the training of these edge-focused models, a working assumption has been that the models will be trained centrally in the cloud, using training data aggregated from several users.

Federated Learning (FL)~\cite{DBLP:conf/aistats/McMahanMRHA17} aims to enable distributed edge devices (or users) to collaboratively \emph{train} a shared prediction model while keeping their personal data private. At a high level, this is achieved by repeating three basic steps: i) local parameters update to a shared prediction model on each edge device, ii) sending the local parameter updates to a central server for aggregation, and iii) receiving the aggregated model back for the next round of local updates. 

A major bottleneck to FL research is the paucity of frameworks that support federated training of workloads on mobile and embedded devices. While several frameworks including Tensorflow Federated~\cite{tensorflowfed,45381} and LEAF~\cite{DBLP:journals/corr/abs-1812-01097} enable simulation of FL clients, they cannot be used to understand the training dynamics and compute the system costs of FL on edge devices. Edge devices exhibit significant heterogeneity in their software stack, compute capabilities, and network bandwidth. All these system-related factors, in combination with the choice of the FL clients and parameter aggregation algorithms, can impact the accuracy and training time of models trained in a federated setting. 

In this paper, we present our exploration of on-device training of FL workloads on Android smartphones and Nvidia Jetson series embedded devices, using the Flower framework~\cite{beutel2020flower}.  Flower offers a stable implementation of the core components of an FL system, and provides higher-level abstractions to enable researchers to experiment and implement new ideas on top of a reliable stack. We first demonstrate how we use the language-, platform- and ML framework-agnostic capabilities of Flower to support on-device training of FL workloads on edge devices with heterogeneous hardware and software stacks. We then deploy these FL clients on various embedded devices as well as on Android smartphones hosted in the Amazon AWS Device Farm (\url{https://aws.amazon.com/device-farm/}). Finally, we present an evaluation to compute various system-related metrics of FL and highlight how this quantification could lead to the design of more efficient FL algorithms.

\section{Related Work}


McMahan et al.~\yrcite{DBLP:conf/aistats/McMahanMRHA17} introduced the basic federated averaging (FedAvg) algorithm and evaluated it in terms of communication efficiency. The optimization of distributed training with and without federated concepts has been covered from many angles \cite{DBLP:journals/corr/abs-1807-11205,DBLP:journals/corr/abs-1810-11787}. Bonawitz et al.~\yrcite{47976} detail the design of a large-scale Google-internal FL system. TFF~\cite{tensorflowfed}, PySyft~\cite{DBLP:journals/corr/abs-1811-04017}, LEAF~\cite{DBLP:journals/corr/abs-1812-01097}, FedML~\cite{chaoyanghe2020fedml} are other open-source frameworks that support research and experimentation of FL workloads. 

Given their relative hardware limitations, most of the works involving machine learning on mobile devices were originally aimed at adapting existing models to specific latency and storage constraints. For this purpose, optimized versions of TensorFlow and PyTorch were developed \cite{david2020tensorflow,NIPS2019_9015} and can now be considered mainstream. Federated Learning, on the other hand,  shifts the training burden from server to the client, which in turn creates the need for developing the adequate supporting back-ends (e.g. back-propagation) for low-power hardware as well. Recent works towards this goal include using low-precision training \cite{NEURIPS2020_13b91943}, controlled updates of biases to reduce memory \cite{NEURIPS2020_81f7acab}, and early-exit models that provide a trade-off between accuracy and compute \cite{10.1145/3446382.3448359}. In time, as hardware capabilities in edge devices improve, we expect to see these advances being deployed by mainstream frameworks.

\section{Primer on Flower} 

\begin{figure}[t]
    \small
    \centering
    \includegraphics[width=0.48\textwidth]{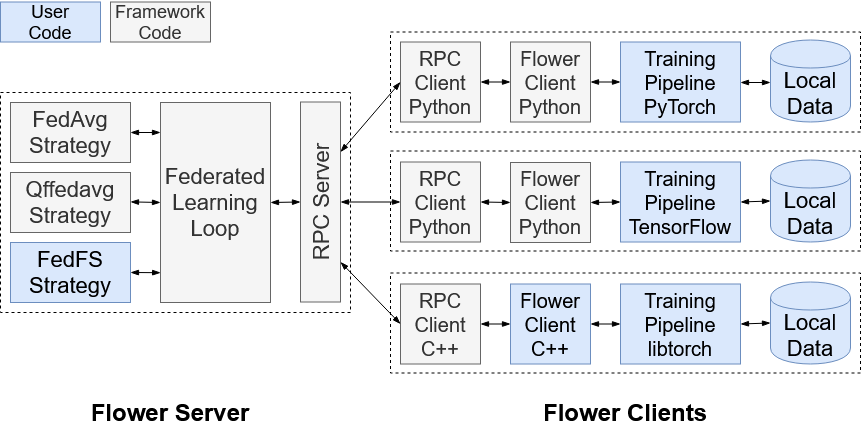}
    \vspace{-0.55cm}
    \caption{Flower framework architecture.\\}
    \vspace{-0.67cm}
    \label{fig:framework-architecture}
\end{figure}

Flower is a novel client-agnostic federated learning framework. One of the underlying design goals of Flower is to enable integrating with an inherently heterogeneous and ever-evolving edge device landscape. There are multiple dimensions of on-device heterogeneity, amongst them are operating systems, machine learning frameworks, programming languages, connectivity, and hardware accelerators.

The Flower core framework, shown in Figure~\ref{fig:framework-architecture}, implements the infrastructure to run these heterogeneous workloads at scale. On the server side, there are three major components involved: the FL loop, the RPC server, and a (user customizable) \emph{Strategy}. Strategy here refers to the federated averaging algorithms (e.g., FedAvg) used for aggregating the model parameters across clients. Clients connect to the RPC server which is responsible for monitoring these connections and for sending and receiving \emph{Flower Protocol} messages. The FL loop is at the heart of the FL process: it orchestrates the learning process and ensures that progress is made. It does not, however, make decisions about \emph{how} to proceed, those decisions are delegated to the currently configured \emph{Strategy} implementation. 

A distinctive property of this architecture is a server which is unaware of the nature of connected clients. This allows to train models across heterogeneous client platforms and implementations, including workloads comprised of different client-side ML frameworks. Furthermore, on-device training and evaluation can be implemented in different programming languages, a property especially important for research on mobile and embedded platforms. These platforms often do not support Python, but rely on specific languages (Java on Android, Swift on iOS) for idiomatic development, or native C/C++ for some embedded devices. Flower achieves a fully language-agnostic interface by offering protocol-level integration. The \emph{Flower Protocol} defines core server-side messages such as \emph{fit} and \emph{evaluate}, which include the (serialized) global model parameters and expect return messages from the client that return either updated model parameters (or gradients) or evaluation results. Each message contains additional user-customizable metadata that allows the server to control on-device hyper-parameters, for example, the number of on-device training epochs. Due to space constraints, we refer the reader to \cite{beutel2020flower} and \url{https://flower.dev/} for more details on Flower.


\section{Flower Clients and on-device training}


In this section, we describe two instances of on-device federated learning with Flower. First, we present how Flower clients can be developed in Java and deployed on Android phones in the AWS Device Farm for federated model training. Next, we discuss the implementation of Flower clients in Python and their deployment on heterogeneous embedded devices such as Nvidia Jetson series and Raspberry Pi.  

\begin{figure}[t]
    \small
    \centering
    \includegraphics[width=0.35\textwidth]{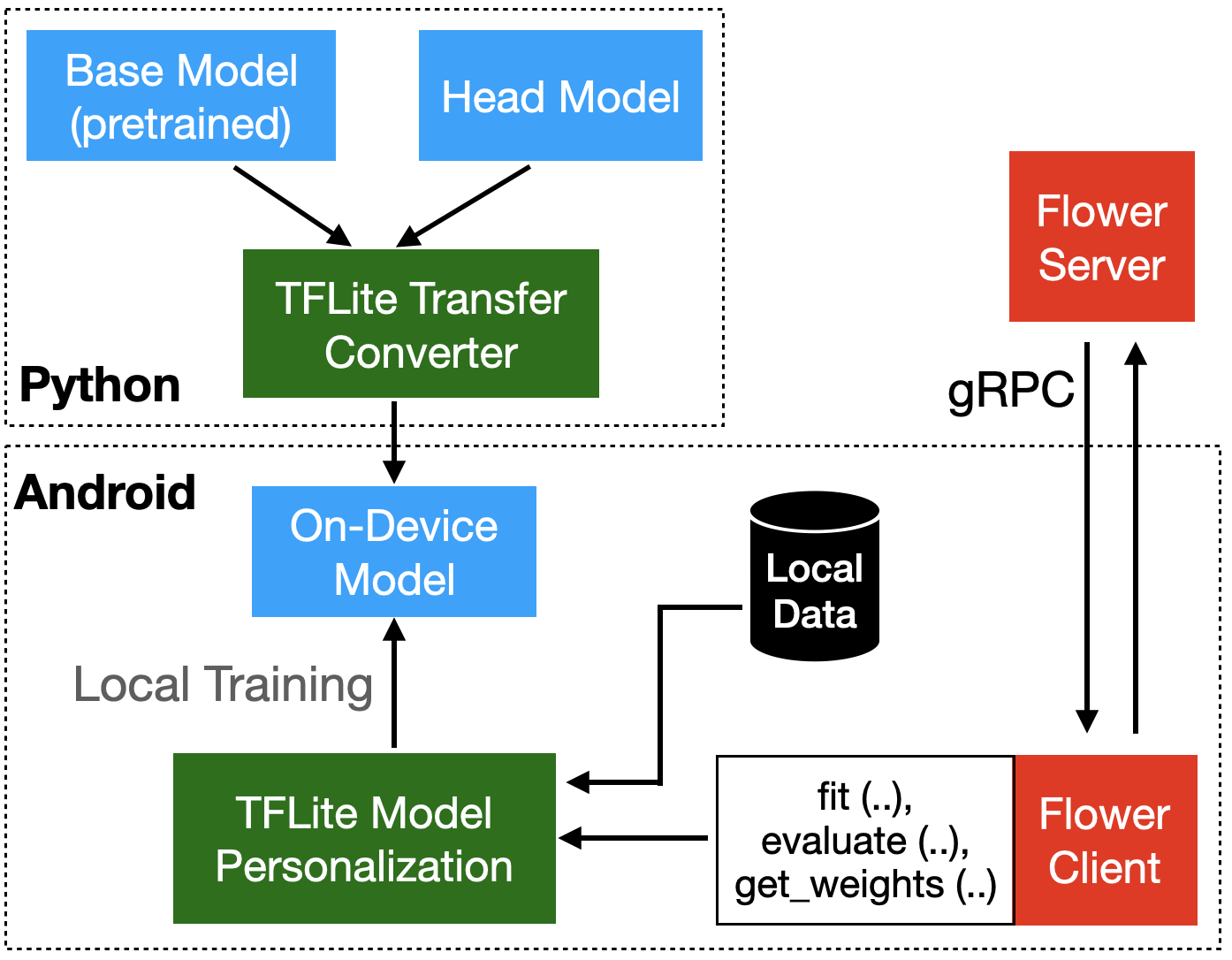}
    \vspace{-0.3cm}
    \caption{Flower Android client architecture.\\}
    \vspace{-0.6cm}
    \label{fig:client-arch}
\end{figure}

\vspace{-0.2cm}
\subsection{Java Flower Clients for Android Smartphones}
\vspace{-0.2cm}
By design, Flower is language-agnostic and can work with any ML framework on the FL client, which maximizes its ability to federate existing training pipelines. However, it also means Flower inherits the limitations of these frameworks that currently offer limited support for on-device training on Android devices.  

In the absence of a full-fledged model training library for Android, we employ the TensorFlow Lite (TFLite) Model Personalization support on Android to perform on-device federated learning. More specifically, as shown in Figure~\ref{fig:client-arch}, we first obtain a pre-trained \emph{Base Model} (e.g., MobileNetV2 without its top layers). The parameters of this model are frozen (i.e., not updated during training), and it is only used as a feature extractor. Next, we define a \emph{Head Model} which corresponds to the task-specific classifier that we want to train using federated learning. The input to the \emph{Head Model} are the features extracted from the \emph{Base Model}, and its weights are randomly initialized. Finally, we use the \textsc{TFLite Transfer Convertor} to port the Base and Head models to TFLite and package them inside an Android application for on-device training. 

Inside the Android application, \textsc{Flower Client} is a class which coordinates with the TFLite Model Personalization libraries and implements the three core methods required for federated training with Flower, namely {(i) get\_weights~(.)}, which gets the current weights of the \emph{Head Model} to support server-side aggregation requests,  (ii) fit~(.), which updates the parameters of the Head Model through local training and (iii) evaluate~(.), which computes test loss on the local dataset and communicates it to the server. 


The \textsc{Flower Client} spawns a background thread and sets up bi-directional streaming RPC with the Flower server using the \textsc{StreamObserver} class. Upon receiving messages from the Flower server, the background thread calls the appropriate TFLite methods to update the parameters of the \emph{Head Model} using the local training data (e.g., by optimizing a cross-entropy loss for classification tasks). The updated parameters are sent to the Flower Server where they are aggregated across all clients, before being sent back to the clients for the next round of training. 


\subsection{Python Flower Clients on Embedded Devices}
Contrary to Android, on-device training on Python-enabled embedded devices can be performed using platform-specific ML frameworks (e.g., TensorFlow, PyTorch). However, one of the open challenges to deploy on-device training workloads arises due to the heterogeneity of computing resources present in such devices. For example, NVIDIA Jetson devices provide GPU acceleration, while Raspberry Pi is limited to CPU-only workloads. Implementing different versions of an FL client for each target platform can quickly become unpractical in real-world scenarios. 

\begin{figure}[t]
    \small
    \centering
    \includegraphics[width=0.4\textwidth]{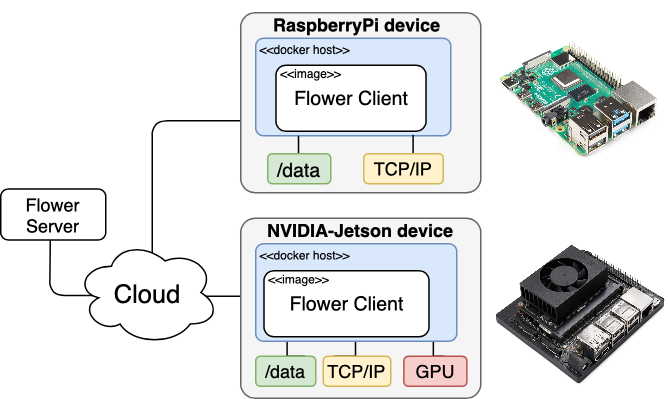}
    \vspace{-0.3cm}
    \caption{Flower clients can easily be deployed to heterogeneous devices by leveraging existing container-based frameworks (e.g. Docker) that interface with the host's hardware.}
    \vspace{-0.5cm}
    \label{fig:embedded_diagram}
\end{figure}

Flower overcomes this challenge by providing a platform-agnostic way of writing FL clients. Figure~\ref{fig:embedded_diagram} illustrates the design of Flower Clients for a NVIDIA Jetson TX2 and a Raspberry Pi device. Despite the differences in these two embedded platforms,  we can use the same code to develop Flower Clients, and run them inside a platform-specific Docker image, enabling the client code to access platform-specific resources (e.g. GPU) when available. 

More importantly, this platform-agnostic capability of Flower can be combined with its language-agnostic capabilities (i.e., support for Python, Java, C++ clients) to enable on-device training of federated workloads in highly-heterogeneous client setups, e.g., when the FL clients include Android phones, Android watches, Raspberry Pis, ARM micro-controllers, Nvidia Jetson devices etc. 

\section{Evaluation}
Our evaluation focuses on quantifying the system costs associated with running FL on various edge devices. In doing so, we also explore how such quantification could help FL developers to design novel algorithms that effectively trade-off between system costs and FL accuracy.


\begin{table}[t]
    \scriptsize
    \caption{Android phones used from the AWS Device Farm}
    \vskip 0.05in
    \centering
    \begin{tabular}{|c|c|c|}
        \toprule
        \textbf{Device Name} & \textbf{Type} & \textbf{OS Version} \\
        \midrule
        Google Pixel 4 & Phone & 10 \\
        Google Pixel 3 & Phone & 10 \\
        Google Pixel 2 & Phone & 9 \\
        Samsung Galaxy Tab S6 & Tablet & 9 \\
        Samsung Galaxy Tab S4 & Tablet & 8.1.0\\
        \bottomrule
    \end{tabular}
    \vspace{-0.5cm}
    \label{tab.devices}
\end{table}

\textbf{Datasets.} Two datasets are used in our evaluation, namely \emph{CIFAR-10} and \emph{Office-31}~\cite{office31}, both of which are examples of object recognition datasets.


\textbf{Deployment Setup.} We run the Flower Server configured with the \texttt{FedAvg} strategy and host it on a cloud virtual machine. Two sets of edge devices are used in the evaluation: Android smartphones and Nvidia Jetson TX2 accelerators. To scale our experiments to a reasonably large number of Android clients with different OS versions, we deploy Flower Clients on the Amazon AWS Device Farm, which enables testing applications on real Android devices accessed through AWS. Table~\ref{tab.devices} list the smartphones from AWS Device Farm used in our evaluation. Nvidia Jetson TX2 devices support full-fledged PyTorch -- this means we could successfully port existing PyTorch training pipelines to implement FL clients on them. 




\begin{table}[t]
\small
\caption{Flower supports implementation of FL clients on any device that has on-device training support. Here we show various FL experiments on Android and Nvidia Jetson devices.}
\vskip 0.1in
\begin{subtable}[h]{0.5\textwidth}
\small
\centering
\begin{tabular}{@{}c|ccc@{}}
\toprule
\textbf{\begin{tabular}[c]{@{}c@{}}Local\\ Epochs (E)\end{tabular}} & \multicolumn{1}{l}{\textbf{Accuracy}} & \textbf{\begin{tabular}[c]{@{}c@{}}Convergence\\ Time (mins)\end{tabular}} & \textbf{\begin{tabular}[c]{@{}c@{}}Energy\\ Consumed (kJ)\end{tabular}}\\ \midrule
1 & 0.48 & 17.63 & 10.21 \\
5 & 0.64 & 36.83 & 50.54 \\
10 & 0.67 & 80.32 & 100.95 \\ \bottomrule
\end{tabular}
\caption{Performance metrics with Nvidia Jetson TX2 as we vary the number of local epochs. Number of clients $C$ is set to 10 and the model is trained for 40 rounds.}
\label{tab:android-a}
\end{subtable}
\hfill
\begin{subtable}[h]{0.5\textwidth}
\small
\centering
\begin{tabular}{@{}c|ccc@{}}
\toprule
\textbf{\begin{tabular}[c]{@{}c@{}}Number\\ of Clients (C)\end{tabular}} & \multicolumn{1}{l}{\textbf{Accuracy}} & \textbf{\begin{tabular}[c]{@{}c@{}}Convergence\\ Time (mins)\end{tabular}} & \textbf{\begin{tabular}[c]{@{}c@{}}Energy\\ Consumed (kJ)\end{tabular}} \\ \midrule
4 & 0.84 & 30.7 & 10.4 \\
7 & 0.85 & 31.3 & 19.72 \\
10 & 0.87 & 31.8 & 28.0 \\ \bottomrule
\end{tabular}
\caption{Performance metrics with Android clients as we vary the number of clients. Local epochs $E$ is fixed to 5 in this experiment and the model is trained for 20 rounds.}
\label{tab:android-b}
\end{subtable}
\label{tab:android}
\vspace{-0.9cm}
\end{table}

\textbf{System Costs of FL.} In Table~\ref{tab:android}, we present various performance metrics obtained on Nvidia TX2 and Android devices. First, we train a ResNet-18 model on the CIFAR-10 dataset on 10 Nvidia TX2 clients. In Table~\ref{tab:android-a}, we vary the number of local training epochs ($E$) performed on each client in a round of FL. Our results show that choosing a higher $E$ results in better FL accuracy, however it also comes at the expense of significant increase in total training time and overall energy consumption across the clients. While the accuracy metrics in Table~\ref{tab:android-a} could have been obtained in a simulated setup, quantifying the training time and energy costs on real clients would not have been possible without a real on-device deployment. As reducing the energy and carbon footprint of training ML models is a major challenge for the community, Flower can assist researchers in choosing an optimal value of $E$ to obtain the best trade-off between accuracy and energy consumption. 

Next, we train a 2-layer DNN classifier (\emph{Head Model}) on top of a pre-trained MobileNetV2 \emph{Base Model} on Android clients for the Office-31 dataset. In Table~\ref{tab:android-b}, we vary the number of Android clients ($C$) participating in FL, while keeping the local training epochs ($E$) on each client fixed to 5. We observe that by increasing the number of clients, we can train a more accurate object recognition model. Intuitively, as more clients participate in the training, the model gets exposed to more diverse training examples, thereby increasing its generalizability to unseen test samples. However, this accuracy gain comes at the expense of high energy consumption -- the more clients we use, the higher the total energy consumption of FL. Again, based on this analysis obtained using Flower, researchers can choose an appropriate number of clients to find a balance between accuracy and energy consumption. 

\begin{table}[]
\small
\caption{Effect of computational heterogeneity on FL training times. Using Flower, we can compute a hardware-specific cutoff $\tau$ (in minutes) for each processor, and find a balance between FL accuracy and training time. $\tau=0$ indicates no cutoff time.}
\centering
\begin{tabular}{@{}cccll@{}}
\toprule
\multicolumn{1}{l}{\textbf{}} & \multicolumn{1}{l}{\textbf{\begin{tabular}[c]{@{}c@{}}GPU \\ ($\tau$ = 0)\end{tabular}}} & \textbf{\begin{tabular}[c]{@{}c@{}}CPU \\ ($\tau$ = 0)\end{tabular}} & \textbf{\begin{tabular}[c]{@{}l@{}}CPU \\ ($\tau$ = 2.23)\end{tabular}} & \textbf{\begin{tabular}[c]{@{}l@{}}CPU \\ ($\tau$ = 1.99)\end{tabular}} \\ \midrule
Accuracy & 0.67 & 0.67 & 0.66 & 0.63 \\
\begin{tabular}[c]{@{}c@{}}Training \\ time (mins)\end{tabular} & 80.32 & \begin{tabular}[c]{@{}c@{}}102\\ (1.27x)\end{tabular} & \begin{tabular}[c]{@{}l@{}}89.15\\ (1.11x)\end{tabular} & \begin{tabular}[c]{@{}l@{}}80.34\\ (1.0x)\end{tabular} \\ \bottomrule
\end{tabular}
\label{tab:compute_hetero}
\vspace{-0.5cm}
\end{table}

\textbf{Computational Heterogeneity across Clients.} FL clients could have vastly different computational capabilities. While newer smartphones are now equipped with GPUs, other devices may have a much less powerful processor. How does this computational heterogeneity impact FL?

For this experiment, we use Nvidia Jetson TX2 as the client device, which has one Pascal GPU and six CPU cores. We repeat the experiment shown in Table~\ref{tab:android-a}, but instead of using the embedded GPU for training, we train the ResNet-18 model on a CPU. In Table~\ref{tab:compute_hetero}, we show that CPU training with local epochs $E=10$ would take $1.27\times$ more time to obtain the same accuracy as the GPU training. This implies that even a single client device with low compute resources (e.g., a CPU) can become a bottleneck and significantly increase the FL training  time. 


Once we obtain this quantification of computational heterogeneity, we can design better federated optimization algorithms. As an example, we implement a modified version of FedAvg where each client device is assigned a cutoff time ($\tau$) after which it must send its model parameters to the server, irrespective of whether it has finished its local epochs or not. This strategy has parallels with the FedProx algorithm \cite{li2018federated} which also accepts partial results from clients. However, the key advantage of using Flower is that we can compute and assign a processor-specific cutoff time for each client. For example, on average it takes 1.99 minutes to complete an FL round on the TX2 GPU. If we set the same time as a cutoff for CPU clients ($\tau = 1.99$ mins) as shown in Table~\ref{tab:compute_hetero}, we obtain the same convergence time as GPU, at the expense of 3\% accuracy drop. With $\tau = 2.23$, a better balance between accuracy and training time could be obtained on a CPU.

\vspace{-0.2cm}
\section{Conclusion}
\vspace{-0.2cm}

We presented our exploration of federated training of models on mobile and embedded devices by leveraging the language-, framework- and platform-agnostic capabilities of the Flower framework. We also presented early results quantifying the system costs of FL on the edge and its implications for the design of more efficient FL algorithms. Although our work is early stage, we hope it will trigger discussions at the workshop related to on-device training for FL and beyond. 

\small
\bibliography{flower_main}

\begin{thebibliography}{22}
\providecommand{\natexlab}[1]{#1}
\providecommand{\url}[1]{\texttt{#1}}
\expandafter\ifx\csname urlstyle\endcsname\relax
  \providecommand{\doi}[1]{doi: #1}\else
  \providecommand{\doi}{doi: \begingroup \urlstyle{rm}\Url}\fi

\bibitem[Abadi et~al.(2016)Abadi, Barham, Chen, Chen, Davis, Dean, Devin,
  Ghemawat, Irving, Isard, Kudlur, Levenberg, Monga, Moore, Murray, Steiner,
  Tucker, Vasudevan, Warden, Wicke, Yu, and Zheng]{45381}
Abadi, M., Barham, P., Chen, J., Chen, Z., Davis, A., Dean, J., Devin, M.,
  Ghemawat, S., Irving, G., Isard, M., Kudlur, M., Levenberg, J., Monga, R.,
  Moore, S., Murray, D.~G., Steiner, B., Tucker, P., Vasudevan, V., Warden, P.,
  Wicke, M., Yu, Y., and Zheng, X.
\newblock Tensorflow: A system for large-scale machine learning.
\newblock In \emph{12th USENIX Symposium on Operating Systems Design and
  Implementation (OSDI 16)}, pp.\  265--283, 2016.

\bibitem[Beutel et~al.(2020)Beutel, Topal, Mathur, Qiu, Parcollet, de~Gusmão,
  and Lane]{beutel2020flower}
Beutel, D.~J., Topal, T., Mathur, A., Qiu, X., Parcollet, T., de~Gusmão, P.
  P.~B., and Lane, N.~D.
\newblock Flower: {A} friendly federated learning research framework.
\newblock \emph{CoRR}, abs/2007.14390, 2020.
\newblock URL \url{https://arxiv.org/abs/2007.14390}.

\bibitem[Bonawitz et~al.(2019)Bonawitz, Eichner, Grieskamp, Huba, Ingerman,
  Ivanov, Kiddon, Konečný, Mazzocchi, McMahan, Overveldt, Petrou, Ramage, and
  Roselander]{47976}
Bonawitz, K., Eichner, H., Grieskamp, W., Huba, D., Ingerman, A., Ivanov, V.,
  Kiddon, C.~M., Konečný, J., Mazzocchi, S., McMahan, B., Overveldt, T.~V.,
  Petrou, D., Ramage, D., and Roselander, J.
\newblock Towards federated learning at scale: System design.
\newblock In \emph{SysML 2019}, 2019.

\bibitem[Cai et~al.(2020)Cai, Gan, Zhu, and Han]{NEURIPS2020_81f7acab}
Cai, H., Gan, C., Zhu, L., and Han, S.
\newblock Tinytl: Reduce memory, not parameters for efficient on-device
  learning.
\newblock In Larochelle, H., Ranzato, M., Hadsell, R., Balcan, M.~F., and Lin,
  H. (eds.), \emph{Advances in Neural Information Processing Systems},
  volume~33, pp.\  11285--11297. Curran Associates, Inc., 2020.

\bibitem[Caldas et~al.(2018)Caldas, Wu, Li, Konecn{\'{y}}, McMahan, Smith, and
  Talwalkar]{DBLP:journals/corr/abs-1812-01097}
Caldas, S., Wu, P., Li, T., Konecn{\'{y}}, J., McMahan, H.~B., Smith, V., and
  Talwalkar, A.
\newblock {LEAF:} {A} benchmark for federated settings.
\newblock \emph{CoRR}, abs/1812.01097, 2018.
\newblock URL \url{http://arxiv.org/abs/1812.01097}.

\bibitem[Chahal et~al.(2018)Chahal, Grover, and
  Dey]{DBLP:journals/corr/abs-1810-11787}
Chahal, K.~S., Grover, M.~S., and Dey, K.
\newblock A hitchhiker's guide on distributed training of deep neural networks.
\newblock \emph{CoRR}, abs/1810.11787, 2018.
\newblock URL \url{http://arxiv.org/abs/1810.11787}.

\bibitem[Chowdhery et~al.(2019)Chowdhery, Warden, Shlens, Howard, and
  Rhodes]{DBLP:journals/corr/abs-1906-05721}
Chowdhery, A., Warden, P., Shlens, J., Howard, A., and Rhodes, R.
\newblock Visual wake words dataset.
\newblock \emph{CoRR}, abs/1906.05721, 2019.
\newblock URL \url{http://arxiv.org/abs/1906.05721}.

\bibitem[David et~al.(2020)David, Duke, Jain, Reddi, Jeffries, Li, Kreeger,
  Nappier, Natraj, Regev, et~al.]{david2020tensorflow}
David, R., Duke, J., Jain, A., Reddi, V.~J., Jeffries, N., Li, J., Kreeger, N.,
  Nappier, I., Natraj, M., Regev, S., et~al.
\newblock Tensorflow lite micro: Embedded machine learning on tinyml systems.
\newblock \emph{arXiv preprint arXiv:2010.08678}, 2020.

\bibitem[Fromm et~al.(2018)Fromm, Patel, and
  Philipose]{10.5555/3327144.3327315}
Fromm, J., Patel, S., and Philipose, M.
\newblock Heterogeneous bitwidth binarization in convolutional neural networks.
\newblock In \emph{Proceedings of the 32nd International Conference on Neural
  Information Processing Systems}, NIPS’18, pp.\  4010–4019, Red Hook, NY,
  USA, 2018. Curran Associates Inc.

\bibitem[Google(2020)]{tensorflowfed}
Google.
\newblock Tensorflow federated: Machine learning on decentralized data.
\newblock \url{https://www.tensorflow.org/federated}, 2020.
\newblock accessed 25-Mar-20.

\bibitem[He et~al.(2020)He, Li, So, Zhang, Wang, Wang, Vepakomma, Singh, Qiu,
  Shen, Zhao, Kang, Liu, Raskar, Yang, Annavaram, and
  Avestimehr]{chaoyanghe2020fedml}
He, C., Li, S., So, J., Zhang, M., Wang, H., Wang, X., Vepakomma, P., Singh,
  A., Qiu, H., Shen, L., Zhao, P., Kang, Y., Liu, Y., Raskar, R., Yang, Q.,
  Annavaram, M., and Avestimehr, S.
\newblock Fedml: A research library and benchmark for federated machine
  learning.
\newblock \emph{arXiv preprint arXiv:2007.13518}, 2020.

\bibitem[Jia et~al.(2018)Jia, Song, He, Wang, Rong, Zhou, Xie, Guo, Yang, Yu,
  Chen, Hu, Shi, and Chu]{DBLP:journals/corr/abs-1807-11205}
Jia, X., Song, S., He, W., Wang, Y., Rong, H., Zhou, F., Xie, L., Guo, Z.,
  Yang, Y., Yu, L., Chen, T., Hu, G., Shi, S., and Chu, X.
\newblock Highly scalable deep learning training system with mixed-precision:
  Training imagenet in four minutes.
\newblock \emph{CoRR}, abs/1807.11205, 2018.
\newblock URL \url{http://arxiv.org/abs/1807.11205}.

\bibitem[Lee et~al.(2019)Lee, Lin, Pushp, Li, Liu, Lee, Xu, Xu, Zhang, and
  Song]{10.1145/3300061.3345447}
Lee, T., Lin, Z., Pushp, S., Li, C., Liu, Y., Lee, Y., Xu, F., Xu, C., Zhang,
  L., and Song, J.
\newblock Occlumency: Privacy-preserving remote deep-learning inference using
  sgx.
\newblock In \emph{The 25th Annual International Conference on Mobile Computing
  and Networking}, MobiCom ’19, New York, NY, USA, 2019. Association for
  Computing Machinery.
\newblock ISBN 9781450361699.
\newblock \doi{10.1145/3300061.3345447}.
\newblock URL \url{https://doi.org/10.1145/3300061.3345447}.

\bibitem[Leontiadis et~al.(2021)Leontiadis, Laskaridis, Venieris, and
  Lane]{10.1145/3446382.3448359}
Leontiadis, I., Laskaridis, S., Venieris, S.~I., and Lane, N.~D.
\newblock It's always personal: Using early exits for efficient on-device cnn
  personalisation.
\newblock In \emph{Proceedings of the 22nd International Workshop on Mobile
  Computing Systems and Applications}, HotMobile '21, pp.\  15–21, New York,
  NY, USA, 2021. Association for Computing Machinery.
\newblock ISBN 9781450383233.
\newblock \doi{10.1145/3446382.3448359}.
\newblock URL \url{https://doi.org/10.1145/3446382.3448359}.

\bibitem[Li et~al.(2018)Li, Sahu, Zaheer, Sanjabi, Talwalkar, and
  Smith]{li2018federated}
Li, T., Sahu, A.~K., Zaheer, M., Sanjabi, M., Talwalkar, A., and Smith, V.
\newblock Federated optimization in heterogeneous networks.
\newblock \emph{arXiv preprint arXiv:1812.06127}, 2018.

\bibitem[Malekzadeh et~al.(2019)Malekzadeh, Athanasakis, Haddadi, and
  Livshits]{malekzadeh2019privacypreserving}
Malekzadeh, M., Athanasakis, D., Haddadi, H., and Livshits, B.
\newblock Privacy-preserving bandits, 2019.

\bibitem[McMahan et~al.(2017)McMahan, Moore, Ramage, Hampson, and
  y~Arcas]{DBLP:conf/aistats/McMahanMRHA17}
McMahan, B., Moore, E., Ramage, D., Hampson, S., and y~Arcas, B.~A.
\newblock Communication-efficient learning of deep networks from decentralized
  data.
\newblock In Singh, A. and Zhu, X.~J. (eds.), \emph{Proceedings of the 20th
  International Conference on Artificial Intelligence and Statistics, {AISTATS}
  2017, 20-22 April 2017, Fort Lauderdale, FL, {USA}}, volume~54 of
  \emph{Proceedings of Machine Learning Research}, pp.\  1273--1282. {PMLR},
  2017.
\newblock URL \url{http://proceedings.mlr.press/v54/mcmahan17a.html}.

\bibitem[Office31(2020)]{office31}
Office31.
\newblock Office 31 dataset.
\newblock \url{https://people.eecs.berkeley.edu/~jhoffman/domainadapt/}, 2020.
\newblock accessed 10-Oct-20.

\bibitem[Paszke et~al.(2019)Paszke, Gross, Massa, Lerer, Bradbury, Chanan,
  Killeen, Lin, Gimelshein, Antiga, Desmaison, Kopf, Yang, DeVito, Raison,
  Tejani, Chilamkurthy, Steiner, Fang, Bai, and Chintala]{NIPS2019_9015}
Paszke, A., Gross, S., Massa, F., Lerer, A., Bradbury, J., Chanan, G., Killeen,
  T., Lin, Z., Gimelshein, N., Antiga, L., Desmaison, A., Kopf, A., Yang, E.,
  DeVito, Z., Raison, M., Tejani, A., Chilamkurthy, S., Steiner, B., Fang, L.,
  Bai, J., and Chintala, S.
\newblock Pytorch: An imperative style, high-performance deep learning library.
\newblock In Wallach, H., Larochelle, H., Beygelzimer, A., d\textquotesingle
  Alch\'{e}-Buc, F., Fox, E., and Garnett, R. (eds.), \emph{Advances in Neural
  Information Processing Systems 32}, pp.\  8026--8037. Curran Associates,
  Inc., 2019.

\bibitem[Ryffel et~al.(2018)Ryffel, Trask, Dahl, Wagner, Mancuso, Rueckert, and
  Passerat{-}Palmbach]{DBLP:journals/corr/abs-1811-04017}
Ryffel, T., Trask, A., Dahl, M., Wagner, B., Mancuso, J., Rueckert, D., and
  Passerat{-}Palmbach, J.
\newblock A generic framework for privacy preserving deep learning.
\newblock \emph{CoRR}, abs/1811.04017, 2018.
\newblock URL \url{http://arxiv.org/abs/1811.04017}.

\bibitem[Sun et~al.(2020)Sun, Wang, Chen, Ni, Agrawal, Cui, Venkataramani,
  El~Maghraoui, Srinivasan, and Gopalakrishnan]{NEURIPS2020_13b91943}
Sun, X., Wang, N., Chen, C.-Y., Ni, J., Agrawal, A., Cui, X., Venkataramani,
  S., El~Maghraoui, K., Srinivasan, V.~V., and Gopalakrishnan, K.
\newblock Ultra-low precision 4-bit training of deep neural networks.
\newblock In Larochelle, H., Ranzato, M., Hadsell, R., Balcan, M.~F., and Lin,
  H. (eds.), \emph{Advances in Neural Information Processing Systems},
  volume~33, pp.\  1796--1807. Curran Associates, Inc., 2020.

\bibitem[Warden \& Situnayake(2019)Warden and Situnayake]{warden2019tinyml}
Warden, P. and Situnayake, D.
\newblock \emph{Tinyml: Machine learning with tensorflow lite on arduino and
  ultra-low-power microcontrollers}.
\newblock " O'Reilly Media, Inc.", 2019.

\end{thebibliography}
\bibliographystyle{mlsys2020}

\end{document}